\documentclass{article}
\usepackage{spconf,amsmath,graphicx}
\usepackage[utf8]{inputenc} 
\usepackage[T1]{fontenc}    
\usepackage{hyperref}       
\usepackage{url}            
\usepackage{booktabs}       
\usepackage{amsfonts}       
\usepackage{nicefrac}       
\usepackage{microtype}      
\usepackage{float}
\usepackage{graphicx}
\usepackage{multirow}

\title{CONTEXTUAL OUT-OF-DOMAIN UTTERANCE HANDLING WITH COUNTERFEIT DATA AUGMENTATION}

\twoauthors
 {Sungjin Lee}
	{Microsoft Research\\  One Microsoft Way\\ Redmond, WA 98052, USA}
 {Igor Shalyminov\sthanks{The work was done during an internship at Microsoft Research}}
	{School of Mathematical and Computer Sciences\\
  Heriot-Watt University\\
  Edinburgh, EH14 4AS, UK}

\begin{document}
%
\maketitle
\begin{abstract}
Neural dialog models often lack robustness to anomalous user input and produce inappropriate responses which leads to frustrating user experience. Although there are a set of prior approaches to out-of-domain (OOD) utterance detection, they share a few restrictions: 
they rely on OOD data or multiple sub-domains, and their OOD detection is context-independent which leads to suboptimal performance in a dialog.
The goal of this paper is to propose a novel OOD detection method that does not require OOD data by utilizing counterfeit OOD turns in the context of a dialog. For the sake of fostering further research, we also release new dialog datasets which are 3 publicly available dialog corpora augmented with OOD turns in a controllable way.
Our method outperforms state-of-the-art dialog models equipped with a conventional OOD detection mechanism by a large margin in the presence of OOD utterances.

\end{abstract}
\section{Introduction}
\label{sec:intro}
Recently, there has been a surge of excitement in developing chatbots for various purposes in research and enterprise. 
Data-driven approaches offered by common bot building platforms (e.g. Google Dialogflow, Amazon Alexa Skills Kit, Microsoft Bot Framework) make it possible for a wide range of users to easily create dialog systems with a limited amount of data in their domain of interest.
Although most task-oriented dialog systems are built for a closed set of target domains, any failure to detect {\em out-of-domain} (OOD) utterances and respond with an appropriate fallback action can lead to frustrating user experience.
There have been a set of prior approaches for OOD detection which require both {\em in-domain} (IND) and OOD data~\cite{nakano2011two,tur2014detecting}.
However, it is a formidable task to collect sufficient data to cover in theory unbounded variety of OOD utterances. In contrast, \cite{lane2007out} introduced an in-domain verification method that requires only IND utterances. Later, with the rise of deep neural networks, \cite{ryu2017neural} proposed an autoencoder-based OOD detection method which surpasses prior approaches without access to OOD data. However, those approaches still have some restrictions such that there must be multiple sub-domains to learn utterance representation and one must set a decision threshold for OOD detection. This can prohibit these methods from being used for most bots that focus on a single task.

The goal of this paper is to propose a novel OOD detection method that does not require OOD data by utilizing counterfeit OOD turns in the context of a dialog. Most prior approaches do not consider dialog context and make predictions for each utterance independently. We will show that this independent decision leads to suboptimal performance even when actual OOD utterances are given to optimize the model and that the use of dialog context helps reduce OOD detection errors.
To consider dialog context, we need to connect the OOD detection task with the overall dialog task. Thus, for this work, we build upon {\em Hybrid Code Networks} (HCN)~\cite{williams2017hybrid} since HCNs achieve state-of-the-art performance in a data-efficient way for task-oriented dialogs, and propose AE-HCNs which extend HCNs with an autoencoder (Figure~\ref{fig:ae-hcn}). 
Furthermore, we release new dialog datasets which are three publicly available dialog corpora augmented with OOD turns in a controlled way (exemplified in Table~\ref{tab:augmentation_example}) to foster further research.~\footnote{\url{https://github.com/sungjinl/icassp2019-ood-dataset.git}}


\begin{table}[t]
  \centering
  \small
    \begin{tabular}{c|l|p{6.7cm}}
      1&\textit{sys:}&Hello, welcome to the Cambridge restaurant system. \\
              &&How may I help you?  \\
      2&\textit{usr:}&\textbf{What time is the next bus from squirrel hill to morewood?} \\
      3&\textit{sys:}&Sorry I didn't catch that. Could you please repeat? \\
      4&\textit{usr:}&\textbf{Bring me to the nearest grocery store}\\
      5&\textit{sys:}&Sorry I didn't catch that. Could you please repeat? \\
      6&\textit{usr:}&\textit{So sorry man} I want a moderately priced restaurant in the west part of town \\
      7&\textit{sys:}&$<$api call$>$
    \end{tabular}
    \caption{Augmented dialog example (OOD utterances in bold and segment-level OOD content in italics.)}
    \label{tab:augmentation_example}
\end{table}

\section{METHODS}
\label{sec:method}
In this section, we first present the standard HCN model. Then we introduce the proposed AE-HCN(-CNN) model, consisting of an autoencoder and a reconstruction score-aware HCN model. Finally, we describe the counterfeit data augmentation method for training the proposed model.

\subsection{HCN}
\begin{figure}[t]
  \centering
  \includegraphics[width=1.0\linewidth]{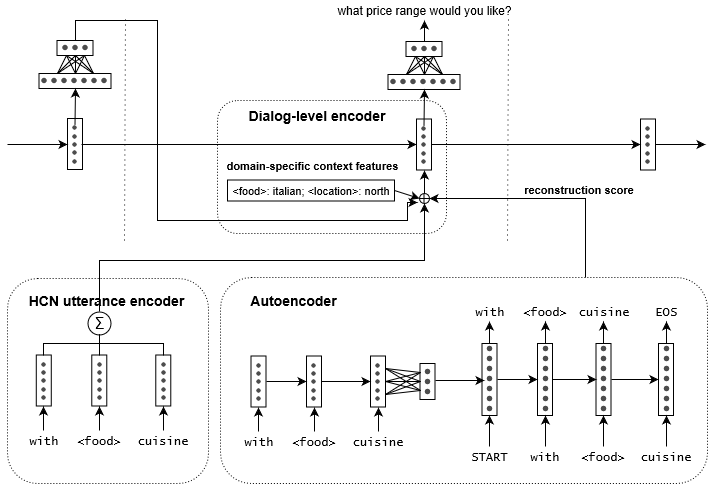}
  \caption{The architecture of AE-HCN which is the same as HCN except for the autoencoder component.}
  \label{fig:ae-hcn}
\end{figure}

As shown in Figure~\ref{fig:ae-hcn}, HCN considers a dialog as a sequence of turns. At each turn, HCN takes a tuple, $(U_t, a_{t-1}, s_t)$, as input to produce the next system action~\footnote{A system action can be either a text output or an api call.} $a_t$, where $U_t$ is a user utterance consisting of $N$ tokens, i.e., $U_t = \{u_1,...,u_N\}$, $a_{t-1}$ a one-hot vector encoding the previous system action and $s_t$ a contextual feature vector generated by domain-specific code.
The user utterance is encoded as a concatenation of a bag-of-words representation and an average of word embeddings of the user utterance:
\begin{equation}
\label{eq:hcn_enc}
    x_t = [bow(U_t); average(e(u_1),...,e(u_N))]
\end{equation}
where $e(\cdot)$ denotes a word embedding layer initialized with GloVe~\cite{pennington2014glove} with 100 dimensions.
HCN then considers the input tuple, $(x_t, a_{t-1}, s_t)$, to update the dialog state through an LSTM~\cite{hochreiter1997long} with 200 hidden units:
\begin{equation}
    h_t = LSTM(h_{t-1}, [x_t;a_{t-1};s_t])
\end{equation}
Finally, a distribution over system actions is calculated by a dense layer with a softmax activation:
\begin{equation}
    P(a_t) = softmax(Wh_t + b)
\end{equation}

\subsection{AE-HCN}
On top of HCN, AE-HCN additionally takes as input an autoencoder's reconstruction score $r_t$ for the user utterance for dialog state update (Figure~\ref{fig:ae-hcn}): 
\begin{equation}
    h_t = LSTM(h_{t-1}, [x_t;a_{t-1};s_t;r_t])
\end{equation}
The autoencoder is a standard seq2seq model which projects a user utterance into a latent vector and reconstructs the user utterance. Specifically, the encoder reads $U_t$ using a GRU~\cite{cho2014properties} to produce a 512-dimensional hidden vector $h_N^{enc} (1 < n < N)$ which in turn gets linearly projected to a 200-dimensional latent vector $z$:
\begin{equation}
    h_n^{enc} = GRU_{enc}(h_{n-1}^{enc}, e(u_n))
\end{equation}
\begin{equation}
    z = W_{z}h_N^{enc} + b_z
\end{equation}
The output of the decoder at step $n$ is a distribution over words:
\begin{equation}
    P_{dec}(y_n) = softmax(W_{dec}h_n^{dec} + b_{dec})
\end{equation}
\begin{equation}
    h_n^{dec} = GRU_{dec}(h_{n-1}^{dec}, e(y_{n-1}))
\end{equation}
\begin{equation}
    h_0^{dec} = W_{dec}z + b_{dec}
\end{equation}
where $GRU_{dec}$ has 512 hidden units. 
The reconstruction score $r_t$ is the normalized generation probability of $U_t$:
\begin{equation}
    r_t = \frac{\sum_{n=0}^{N}\log P_{dec}(u_n)}{N}
\end{equation}
 
\subsection{AE-HCN-CNN}
AE-HCN-CNN is a variant of AE-HCN where user utterances are encoded using a CNN layer with max-pooling (following~\cite{kim2014convolutional}) rather than equation~\ref{eq:hcn_enc}:
\begin{equation}
    x_t = Pooling_{max}(CNN(e(u_1),...,e(u_n)))
\end{equation}
The CNN layer considers two kernel sizes (2 and 3) and has 100 filters for each kernel size.

\subsection{Counterfeit Data Augmentation}
To endow an AE-HCN(-CNN) model with a capability of detecting OOD utterances and producing fallback actions without requiring real OOD data, we augment training data with counterfeit turns.
We first select arbitrary turns in a dialog at random according to a counterfeit OOD probability $\rho$, and insert counterfeit turns before the selected turns.
A counterfeit turn consists of a tuple $(U_t, a_{t-1}, s_t, r_t)$ as input and a fallback action $a_t$ as output.
We copy $a_{t-1}$ and $s_t$ of each selected turn to the corresponding counterfeit turns 
since OOD utterances do not affect previous system action and feature vectors generated by domain-specific code.
Now we generate a counterfeit $U_t$ and $r_t$. Since we don't know OOD utterances a priori, we randomly choose one of the user utterances of the same dialog to be $U_t$. This helps the model learn to detect OOD utterances because a random user utterance is contextually inappropriate just like OOD utterances are. We generate $r_t$ by drawing a sample from a uniform distribution, $U[\alpha, \beta]$, where $\alpha$ is the maximum reconstruction score of training data and $\beta$ is an arbitrary large number. The rationale is that the reconstruction scores of OOD utterances are likely to be larger than $\alpha$ but we don't know what distribution the reconstruction scores of OOD turns would follow. Thus we choose the most uninformed distribution, i.e., a uniform distribution so that the model may be encouraged to consider not only reconstruction score but also other contextual features such as the appropriateness of the user utterance given the context, changes in the domain-specific feature vector, and what action the system previously took.

\section{DATASETS}
\label{sec:data}
To study the effect of OOD input on dialog system's performance, we use three task-oriented dialog datasets: bAbI6~\cite{bordes2016learning} initially collected for Dialog State Tracking Challenge 2~\cite{DBLP:conf/sigdial/HendersonTW14}; GR and GM taken from Google multi-domain dialog datasets ~\cite{shah2018building}. Basic statistics of the datasets are shown in Table~\ref{tab:data_stats}. 
bAbI6 deals with restaurant finding tasks, GM buying a movie ticket, and GR reserving a restaurant table, respectively. We generated distinct action templates by replacing entities with slot types and consolidating based on dialog act annotations.

We augment test datasets (denoted as Test-OOD in Table~\ref{tab:data_stats}) with real user utterances from other domains in a controlled way. Our OOD augmentations are as follows:
\begin{itemize}
\small
  \item \textit{OOD utterances}: user requests from a foreign domain~--- the desired system behavior for such input is a fallback action,
  \item \textit{segment-level OOD content}: interjections in the user in-domain requests~--- treated as valid user input and is supposed to be handled by the system in a regular way.
\end{itemize}

These two augmentation types reflect a specific dialog pattern of interest (see Table \ref{tab:augmentation_example}): first, the user utters a request from another domain at an arbitrary point in the dialog (each turn is augmented with the probability $p_{ood\_start}$, which is set to 0.2 for this study), and the system answers accordingly. This may go on for several turns in a row~---each following turn is augmented with the probability $p_{ood\_cont}$, which is set to 0.4 for this study. Eventually, the OOD sequence ends up and the dialog continues as usual, with a segment-level OOD content of the user affirming their mistake.
While we introduce the OOD augmentations in a controlled programmatic way, the actual OOD content is natural. The OOD utterances are taken from dialog datasets in several foreign domains: 1) Frames dataset~\cite{DBLP:conf/sigdial/AsriSSZHFMS17}~--- travel booking (1198 utterances); 2) Stanford Key-Value Retrieval Network Dataset~\cite{DBLP:conf/sigdial/EricKCM17}~--- calendar scheduling, weather information retrieval, city navigation (3030 utterances); 3) Dialog State Tracking Challenge 1~\cite{DBLP:conf/sigdial/WilliamsRRB13}~--- bus information (968 utterances).

In order to avoid incomplete/elliptical phrases, we only took the first user's utterances from the dialogs.
For segment-level OOD content, we mined utterances with the explicit affirmation of a mistake from Twitter and Reddit conversations datasets~--- 701 and 500 utterances respectively. 

\begin{table}[h]
\small
\centering
\begin{tabular}{|l|c|c|c|c|c|}
\hline
\bf{bAbI6}	& \bf{Train}	& \bf{Dev}	& \bf{Test}	& \bf{Test-OOD}	\\ \hline
\# dialogs	& 1618	& 500	& 1117	& 1117 \\ 
Avg. turns per dialog	& 20.08	& 19.30	& 22.07	& 27.27	\\ \hline
\bf{GR}	& \bf{Train}	& \bf{Dev}	& \bf{Test}	& \bf{Test-OOD}	\\ \hline
\# dialogs	& 1116	& 349	& 775	& 775	\\ 
Avg. turns per dialog	& 9.07	& 6.53	& 6.87	& 9.01	\\ \hline
\bf{GM}	& \bf{Train}	& \bf{Dev}	& \bf{Test}	& \bf{Test-OOD}	\\ \hline
\# dialogs	& 362	& 111	& 252	& 252	\\ 
Avg. turns per dialog	& 8.78	& 9.14	& 8.73	& 11.25	\\ \hline
\end{tabular}
\caption{Data statistics. The numbers of distinct system actions are 58, 247, and 194 for bAbI6, GR, and GM, respectively.}
\label{tab:data_stats}
\end{table}

\section{EXPERIMENTAL SETUP AND EVALUATION}
\label{sec:evaluation}

\begin{table*}[t]
\center
\small
\begin{tabular}{|l|c|c|c|c|c|c|c|c|c|}
  \hline
\bf{Domain} & \multicolumn{3}{|c|}{\bf{bAbI6}} & \multicolumn{3}{|c|}{\bf{GR}} & \multicolumn{3}{|c|}{\bf{GM}} \\  
\hline
\bf{Test Data} & \bf{Test} & \multicolumn{2}{|c|}{\bf{Test-OOD}} & \bf{Test} & \multicolumn{2}{|c|}{\bf{Test-OOD}} & \bf{Test} & \multicolumn{2}{|c|}{\bf{Test-OOD}} \\
\hline
\bf{Metrics} & \bf{P@1} & \bf{P@1} & \bf{OOD F1} & \bf{P@3} & \bf{P@3} & \bf{OOD F1} & \bf{P@3} & \bf{P@3} & \bf{OOD F1}  \\
\hline
HCN & 53.41 & 41.95 & 0 & \bf{58.89} & 41.65 & 0 & 41.18 & 27.08 & 0 \\
AE-HCN-Indep & 31.29 & 41.06 & 48.68 & 51.90 & 55.42 & 71.52 & 31.12 & 42.78 & 64.35 \\
AE-HCN  & 53.58 & 55.04 & \bf{73.41} & 56.97 & 58.90 & 74.67 & 40.61 & 48.59 & \bf{69.31} \\
AE-HCN-CNN & \bf{55.04} & \bf{55.35} & 70.38 & 58.32 & \bf{64.51} & \bf{81.33} & \bf{45.12} & \bf{52.79} & 68.59 \\
\hline
\end{tabular}
\caption{Evaluation results. P@K means Precision@K. OOD F1 denotes f1-score for OOD detection over utterances.}
\label{tab:res}
\end{table*}

We comparatively evaluate four different models: 1) an HCN model trained on in-domain training data; 2) an AE-HCN-Indep model which is the same as the HCN model except that it deals with OOD utterances using an independent autoencoder-based rule to mimic~\cite{ryu2017neural} -- when the reconstruction score is greater than a threshold, the fallback action is chosen; we set the threshold to the maximum reconstruction score of training data; 3) an AE-HCN(-CNN) model trained on training data augmented with counterfeit OOD turns -- the counterfeit OOD probability $\rho$ is set to 15\% and $\beta$ to 30.
We apply dropout to the user utterance encoding with the probability 0.3. We use the Adam optimizer~\cite{kingma2014adam}, with gradients computed on mini-batches of size 1 and clipped with norm value 5. The learning rate was set to $1 \times 10^{-3}$ throughout the training and all the other hyperparameters were left as suggested in \cite{kingma2014adam}. We performed early stopping based on the performance of the evaluation data to avoid overfitting. We first pretrain the autoencoder on in-domain training data and keep it fixed while training other components.

\begin{figure}[htb]
\center
\includegraphics[width=0.8\linewidth]{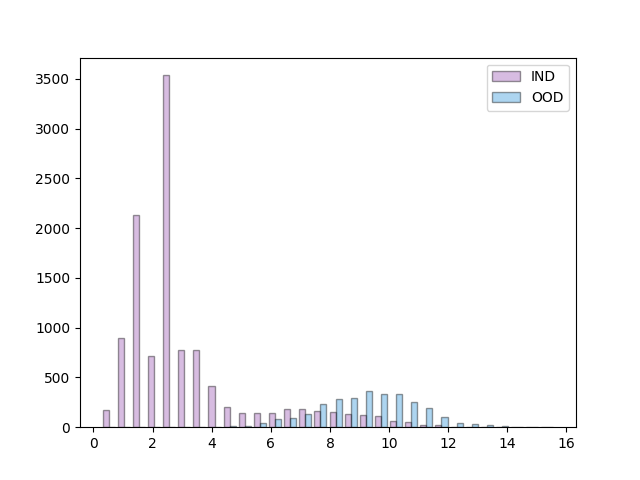}
\caption{Histograms of AE reconstruction scores for the bAbI6 test data. The histograms for other datasets follow similar trends.}
\label{fig:vis}
\end{figure}

The result is shown in Table~\ref{tab:res}. Since there are multiple actions that are appropriate for a given dialog context, we use per-utterance {\em Precision@K} as performance metric. We also report f1-score for OOD detection to measure the balance between precision and recall.  
The performances of HCN on Test-OOD are about 15 points down on average from those on Test, showing the detrimental impact of OOD utterances to such models only trained on in-domain training data. 
AE-HCN(-CNN) outperforms HCN on Test-OOD by a large margin about 17(20) points on average while keeping the minimum performance trade-off compared to Test. Interestingly, AE-HCN-CNN has even better performance than HCN on Test, indicating that, with the CNN encoder, counterfeit OOD augmentation acts as an effective regularization.
In contrast, AE-HCN-Indep failed to robustly detect OOD utterances, resulting in much lower numbers for both metrics on Test-OOD as well as hurting the performance on Test. This result indicates two crucial points: 1) the inherent difficulty of finding an appropriate threshold value without actually seeing OOD data; 2) the limitation of the models which do not consider context. For the first point, Figure~\ref{fig:vis} plots histograms of reconstruction scores for IND and OOD utterances of bAbI6 Test-OOD. If OOD utterances had been known a priori, the threshold should have been set to a much higher value than the maximum reconstruction score of IND training data (6.16 in this case).

For the second point, Table~\ref{tab:thresh} shows the search for the best threshold value for AE-HCN-Indep on the bAbI6 task when given actual OOD utterances (which is highly unrealistic for the real-world scenario). Note that the best performance achieved at 9 is still not as good as that of AE-HCN(-CNN). This implies that we can perform better OOD detection by jointly considering other context features.

\begin{table}[h!]
\small
\center
\begin{tabular}{|c|c|c|}
  \hline
\bf{Threshold} & \bf{Precision@1} & \bf{OOD F1} \\
\hline
6 & 40.39 & 48.38 \\
7 & 42.56 & 50.46 \\
8 & 43.69 & 51.08 \\
9 & 52.21 & 63.86 \\
10 & 47.27 & 44.44 \\
\hline
\end{tabular}
\caption{Performances of AE-HCN-Indep on bAbI6 Test-OOD with different thresholds.}
\label{tab:thresh}
\end{table}

Finally, we conduct a sensitivity analysis by varying counterfeit OOD probabilities.
Table~\ref{tab:rate} shows performances of AE-HCN-CNN on bAbI6 Test-OOD with different $\rho$ values, ranging from 5\% to 30\%. The result indicates that our method manages to produce good performance without regard to the $\rho$ value.
This superior stability nicely contrasts with the high sensitivity of AE-HCN-Indep with regard to threshold values as shown in Table~\ref{tab:thresh}.

\begin{table}[h!]
\small
\center
\begin{tabular}{|c|c|c|c|}
\hline
\bf{Test Data} & \bf{Test} & \multicolumn{2}{|c|}{\bf{Test-OOD}} \\
\hline
\bf{Counterfeit} & \multirow{2}{*}{\bf{Precision@1}} & \multirow{2}{*}{\bf{Precision@1}} & \multirow{2}{*}{\bf{OOD F1}} \\
\bf{OOD Rate} & & & \\
\hline
5\% & 55.25 & 55.48 & 69.72 \\
10\% & 55.08 & 57.29 & 74.73 \\
15\% & 55.04 & 55.35 & 70.38 \\
20\% & 53.48 & 56.53 & 75.55 \\
25\% & 53.72 & 56.66 & 73.13 \\
30\% & 54.87 & 56.02 & 71.44 \\
\hline
\end{tabular}
\caption{Performances of AE-HCN-CNN on bAbI6 Test-OOD with varying counterfeit OOD rates.}
\label{tab:rate}
\end{table}

\section{CONCLUSION}
\label{sec:conclusion}
We proposed a novel OOD detection method that does not require OOD data without any restrictions by utilizing counterfeit OOD turns in the context of a dialog. We also release new dialog datasets which are three publicly available dialog corpora augmented with natural OOD turns to foster further research.
In the presence of OOD utterances, our method outperforms state-of-the-art dialog models equipped with an OOD detection mechanism by a large margin~--- more than 17 points in Precision@K on average~--- while minimizing performance trade-off on in-domain test data.
The detailed analysis sheds light on the difficulty of optimizing context-independent OOD detection and justifies the necessity of context-aware OOD handling models. 
We plan to explore other ways of scoring OOD utterances than autoencoders. For example, variational autoencoders or generative adversarial networks have great potential. We are also interested in using generative models to produce more realistic counterfeit user utterances.


\bibliographystyle{IEEEbib}
\bibliography{refs}

\end{document}